\documentclass[sigconf]{aamas} 
\renewcommand\footnotetextcopyrightpermission[1]{} %
\usepackage{balance} %
\usepackage{amsmath}
\newcommand{\argmax}{\operatorname*{argmax}}

\newcommand{\method}{\text{SmtW}}

\DeclareMathOperator{\States}{S}
\DeclareMathOperator{\Memory}{H}
\DeclareMathOperator{\Actions}{A}
\DeclareMathOperator{\Kernel}{P}
\DeclareMathOperator{\R}{R}
\DeclareMathOperator{\B}{B}
\DeclareMathOperator{\sm}{softmax}

\usepackage{wrapfig}
\newcommand{\overbar}[1]{\mkern 1.5mu\overline{\mkern-1.5mu#1\mkern-1.5mu}\mkern 1.5mu}

\usepackage{mathtools}
\mathtoolsset{showonlyrefs=true}

\setcopyright{ifaamas}
\acmConference[AAMAS '21]{Proc.\@ of the 20th International Conference on Autonomous Agents and Multiagent Systems (AAMAS 2021)}{May 3--7, 2021}{Online}{U.~Endriss, A.~Now\'{e}, F.~Dignum, A.~Lomuscio (eds.)}
\copyrightyear{2021}
\acmYear{2021}
\acmDOI{}
\acmPrice{}
\acmISBN{}

\acmSubmissionID{210}

\title[Show me the Way]{Show me the Way:\\Intrinsic Motivation from Demonstrations}

\author{L\'eonard Hussenot}
\affiliation{
  \institution{Google Research, Brain Team\\Univ. Lille, CNRS, Inria Scool, UMR 9189 CRIStAL}}

\author{Robert Dadashi}
\affiliation{
  \institution{Google Research, Brain Team}}

\author{Matthieu Geist}
\affiliation{
  \institution{Google Research, Brain Team}}
  
  \author{Olivier Pietquin}
\affiliation{
  \institution{Google Research, Brain Team}}

\begin{abstract}

The study of exploration in the domain of decision making has a long history but remains actively debated. From the vast literature that addressed this topic for decades under various points of view (\textit{e.g.}, developmental psychology, experimental design, artificial intelligence), intrinsic motivation emerged as a concept that can practically be transferred to artificial agents. Especially, in the recent field of Deep Reinforcement Learning (RL), agents implement such a concept (mainly using a novelty argument) in the shape of an exploration bonus, added to the task reward, that encourages visiting the whole environment. This approach is supported by the large amount of theory on RL for which convergence to optimality assumes exhaustive exploration. Yet, Human Beings and mammals do not exhaustively explore the world and their motivation is not only based on novelty but also on various other factors (\textit{e.g.}, curiosity, fun, style, pleasure, safety, competition, \textit{etc.}). They optimize for life-long learning and train to learn transferable skills in playgrounds without obvious goals. They also apply innate or learned priors to save time and stay safe. For these reasons, we propose to learn an exploration bonus from demonstrations that could transfer these motivations to an artificial agent with little assumptions about their rationale. Using an inverse RL approach, we show that complex exploration behaviors, reflecting different motivations, can be learnt and efficiently used by RL agents to solve tasks for which exhaustive exploration is prohibitive. 
\end{abstract}

\newcommand{\BibTeX}{\rm B\kern-.05em{\sc i\kern-.025em b}\kern-.08em\TeX}

\makeatletter
\def\@copyrightspace{\relax}
\makeatother
\begin{document}

\pagestyle{fancy}
\fancyhead{}

\maketitle

\section{Introduction}
Intrinsic motivation \citep{deci2010intrinsic} has emerged as one explanation for humans' and animals' impressive learning capabilities. Steered by the need to explore their environment (whether this need is satiable or not has been a fierce debate in the behavioral psychological community \citep{hunt1965intrinsic}), they are able to discover near-optimal strategies in very efficient ways. Designing artificial agents presenting such capabilities is a central goal of modern artificial intelligence and Reinforcement Learning (RL) is a popular candidate to do so. 
RL has addressed a variety of sequential-decision-making problems whether in games~\citep{tesauro1995temporal, mnih2015human, silver2016mastering} or robotics~\citep{abbeel2004apprenticeship, andrychowicz2020learning}. Nevertheless, some simple problems remain unsolved. Current state-of-the-art methods struggle to find good policies in environments (1) where constant negative rewards may discourage the agent to explore (\textit{e.g.}, the \textit{Pitfall!} game from Atari), (2) where the reward is so sparse that an agent does not find any (\textit{e.g.}, the \textit{Montezuma's Revenge} Atari game), (3) where state and action space are big (\textit{e.g.}, text worlds). These tasks remain fairly easy for humans, though.
In order to tackle these specific problems, the use of reward bonuses, inspired by animal curiosity, was proposed to steer the agent's exploration~ \citep{csimcsek2006intrinsic, strehl2008analysis}. Even though different intrinsic bonuses have been proposed, a large majority rely on the same principle: reward for \textit{novelty}. These methods mostly differ in how they compute this notion of \textit{newness}.  Count-based methods do it by counting how often the agent has encountered a given state~ \citep{strehl2008analysis}. Pseudo-counts methods~ \citep{bellemare2016unifying, ostrovski2017count} allow to approximate counts in large state spaces. Prediction error is also used to measure novelty, either by computing the agent's ability to predict the future~\citep{pathak2017curiosity} or random statistics about the current state~\citep{burda2018exploration}. Some restrict novelty to state-action pairs that have an impact on the agent~\citep{raileanu2020ride} or derive empowerment metrics~\citep{mohamed2015variational} using mutual information. All these methods naturally encourage the discovery of new states through exhaustive exploration. Yet, in most realistic environments, exhaustive exploration is (1) not feasible due to the size of the state-action space, (2) not desirable as most behaviors are unlikely to be relevant for the task at hand.

Nonetheless, human and more generally mammals exploration behaviors are governed by various motivations and constraints. Intelligent Beings do not have unlimited resources of time and energy. They optimize these resources to survive and reproduce but also to have fun~\citep{holloway2004children}, to help others~\citep{byrne1990machiavellian} or to satisfy their curiosity. \citet{oudeyer2009intrinsic} make the difference between homeostatic motivations (that encourage to stay in the ``comfort zone'' and generally correspond to desires that can be satiated) and heterostatic motivations (that push organisms out of equilibrium but cannot be satiated). These many desires shape the way organisms interact with their environment, encouraging them to discover new things but also to protect themselves, avoiding over-surprising events with mechanisms like fear~\citep{lang2000fear}. \citet{berseth2019smirl} exemplified how to exploit such priors by implementing a ``homeostasis'' objective for RL, thereby showing how different from ``novelty seeking'' these priors can be. Eventually, the resource constraints stop organisms from exploring exhaustively their environment and push them to transfer knowledge from past experience. \citet{hunt1965intrinsic} developed the idea of optimal incongruity: high-novelty is not rewarded as much as intermediate-level novelty, suggesting how curiosity is tightly connected to fear. More recently, \citet{kidd2012goldilocks} supported this hypothesis with experiments on children curiosity.  Overall, novelty methods fail to model correctly human curiosity as they consider that ``the newer, the better''. This failure calls for a new way of definingour agent's intrinsic motivations.

In an arbitrary environment, exhaustive exploration is desirable and leads to convergence with theoretical guarantees~\citep{strehl2009reinforcement}. But when the exploration presents some structure, one can transfer skills and priors from similar environments.
\citet{dubey2018investigating} exemplified, in the case of simple video games, how humans priors help us to solve new problems. The authors enlighten how humans struggle to play the same underlying video game with change of the object semantics, physics modifications (\textit{e.g.}, the gravity is rotated) or with visual similarities transformations. Overall, they show how much of the human's ability to solve a new game in a zero-shot manner is due to their prior on the environment.

In this paper, instead of hard coding what we think an agent's motivations should be (e.g. novelty), we propose to learn a bonus that captures these sources of motivation from demonstrations.
By adopting this approach, we expect to learn a bonus that implicitly helps reproducing a structured exploration behavior (\textit{i.e.} using priors from the demonstrations to reduce the search space), in lieu of an exhaustive one. We also argue that, to a certain extent at least, this can happen without the need of extra modelling inspired by cognitive or behavioral research.
 To do so, we cast this problem as an inverse RL problem with the difference that only some fraction of the reward optimized by an observed agent is hidden: the intrinsic motivation bonus. The task-related reward remains provided by the environment. We then build upon \citet{klein2013cascaded} to propose a method that allows us to recover the intrinsic motivation from demonstrations.

Therefore, our contributions are the following: 
\begin{enumerate}
    \item a modelling that allows for disentangling the reward optimized by a demonstrator from its intrinsic motivation bonus; 
    \item an architecture, that we call ``\textit{Show me the Way}'' (\method{}), based on a cascade of supervised learning methods that extracts that exploration bonus from demonstrations; 
    
    \item an empirical evaluation showing that SmtW is able to capture different exploration priors explained by various types of motivations.
\end{enumerate}
To evaluate \method{}, we validate a set of hypotheses on a controlled environment. We notably find that our method can learn structures and styles, transfer useful priors and encourages long-term planning.

\section{Background}

\textbf{Markov Decision Processes.} In Reinforcement Learning (RL), an agent learns to behave optimally through interactions with an environment. This is usually formalized as a Markov Decision Process (MDP)~\citep{sutton2018reinforcement,puterman2014markov}, a tuple $(\States, \Actions, \Kernel, \R,\gamma)$ with $\States$ the set  of states, $\Actions$ the set of actions (assumed discrete here), $\Kernel: \States \times \Actions \to \textit{P}(\States)$ the Markovian transition kernel defining the dynamic of the environment, $\R:\States \times \Actions \to \mathbb{R}$ a bounded reward function and $\gamma \in \left[0, 1\right[$ a discount factor. The agent interacts with the environment through a (here deterministic) policy $\pi:\States \to \Actions$. The quality of a given policy is quantified by the associated state-action value function, or $Q$-function. It is the expected discounted cumulative reward for starting from $s$, taking action $a$, and following $\pi$ afterward: $Q^{\pi}(s,a)=\mathbb{E}_\pi[\sum_{t\geq 0}\gamma^t r_t|s_0=s, a_0=a]$,
with $a_t = \pi(s_t)$, $r_t = \R(s_t, a_t)$ and $s_{t+1} \sim \Kernel(.|s_t, a_t)$.
By construction, it satisfies the Bellman equation: for any $s,a$, $Q^{\pi}(s,a) = \R(s,a) + \gamma\sum_{s'} \Kernel(s'|s,a)  Q^{\pi}(s', \pi(s'))$.
An optimal policy $\pi^*$ satisfies component-wise $Q^{\pi_*}\geq Q^\pi$, for any policy $\pi$. Let $Q^*=Q^{\pi_*}$ be the associated (unique) optimal $Q$-function, any deterministic optimal policy is greedy with respect to it: $\pi^*(s) \in \argmax_a Q^*(s,a)$.

\textbf{Exploration Bonus.} A common strategy to encourage exploration is to augment the reward function with a bonus. This bonus generally depends on past history. For example, a bonus rewarding novelty requires remembering what has been experienced so far. Write $h_t=(s_0,a_0,\dots,s_{t-1},a_{t-1},s_t)$ the history up to time $t$, and $\Memory$ the set of all histories. Generally speaking, we abstract a bonus as $\B:\Memory\times \Actions \rightarrow \mathbb{R}$, and use it for addressing the dilemma between exploration and exploitation, which thus amounts for the agent to optimize for $R(s_t,a_t) + B(h_t,a_t)$ instead of simply $R(s_t,a_t)$.
This state-of-the-art exploration bonuses all rely on memory, \textit{e.g.} by counting the state visitation~ \citep{strehl2008analysis} or through updates of a neural networks~ \citep{burda2018exploration, ostrovski2017count}.
These exploration bonuses are designed to express the prior that any source of novelty is good for exploration.

Such a prior on what is good for exploration is task-specific. \citet{taiga2020bonus} showed that state-of-the-art bonuses were degrading performances in most Atari games. 
\section{Show me the Way}
Rather than handcrafting a bonus that encodes what we think intrinsic motivation should be (\textit{e.g.} using novelty), we propose to learn it from  demonstrations of exploratory behaviours.

We thus assume that the demonstrator learns to solve a task by exploring its environment and a simple solution would be to perform behavior cloning. Because the demonstrator is likely to use past interactions to make decisions (remembering what has been already tried so far), we could frame our problem as learning, in a supervised manner, a mapping from histories to actions. Yet, behavioral cloning suffers the behavioral drift \citep{ross2010efficient}, which would be exacerbated in the case of history dependent policies . Moreover, we would like to transfer this behavior to new environments and possibly to new tasks.

Imitation learning classically assumes that experts are optimizing an MDP with an unknown reward function $\R_E(s,a)$. Note that this introduces a modelling bias, \textit{i.e.} a human performing a task is not necessarily explicitly solving an MDP. In this study, we do not tackle the standard problem of inferring an optimal behavior from demonstrations, but of estimating an exploratory behavior. Yet, the latter can be reduced to the former. Taking inspiration from RL, and especially from the body of work about the exploration-exploitation dilemma, we assume that our demonstrator is optimizing for an unknown reward function $\R_E(s,a)$, standing for the task objective, augmented with a trajectory-dependent intrinsic bonus $\B_E(h,a)$, standing for how the environment is explored. Making this bonus depend on past interactions is important as we can reasonably assume that exploration is based on memory (one would not try to always reproduce situations that were already seen). Then, we assume that the expert is optimal for the bonus-augmented reward $\R_E(s,a) + \B_E(h,a)$, in the augmented MDP $\{\Memory, A, P, \R_E+\B_E, \gamma\}$. That is the original MDP, with the state space being replaced by the set of all trajectories and the reward function being augmented with a bonus function.
So, we reduced our original problem to Inverse Reinforcement Learning (IRL): learn a function $\widehat{RB}:\Memory\times \Actions \rightarrow \mathbb{R}$ such that the demonstrator is the (unique) optimal policy. By design, $\widehat{RB} = \R_E + \B_E$ is a solution to this problem (even if it is not learnable exactly, as the optimal policy is invariant to many reward transformations \citep{ng1999policy}). Yet, we also assume that we know the task's reward $R$, or at least that we observe it in the demonstrations. Formally, it may be different from the reward $R_E$ (even if we assume that it leads to a similar optimal behavior), but we can leverage it to disentangle the task contribution and the exploration one. For doing so, we propose to learn a bonus function $\hat{B}:\mathcal{H}\times A \rightarrow \mathbb{R}$ such that the demonstrator is optimal for $R + \hat{B}$. Notice that it does not change the problem, as it is just a reparameterization of the previous one (by setting $\hat{B} = \widehat{RB} - \R$).

Thus, our IRL problem has additional constraints. First, we want to recover the bonus, for transferring to new environments or new tasks, this preclude using imitation learning methods that do not explicitly recover rewards (IRL is mandatory). Second, our specific parameterization ($\R+\B$) precludes using IRL methods that would not allow using the observation of the task reward $\R$ along the expert trajectories. Third, the function we want to estimate is history-dependent which requires an IRL method able to use sequences as inputs.

\textbf{Formalism.}
We assume to have access to demonstrations that are optimal according to the (known) reward of the environment \emph{plus} an (unknown) intrinsic bonus. The environment being assumed Markovian, knowing the current state is enough to act optimally according to the task (optimizing for the environment's reward). Yet, the demonstrator also optimizes its exploration bonus, that depends on the past. To formalize things, we consider that the demonstrations are provided by a policy $\pi_E:\Memory\rightarrow\Actions$, and that the policy is optimal for the augmented MDP $(\Memory, \Actions, \Kernel, \R_E+\B_E)$, where $\Memory$ replaces $\States$ and $\R_E+\B_E$ replaces $\R$.

We frame our problem as learning the bonus $\B_E$ from trajectories sampled from $\pi_E$.

\textbf{Our approach.}
If we cannot naively apply any existing IRL algorithm to our problem, it can be a source of inspiration. Especially, one suits well our problem: the set-policy framework \citep{piot2016bridging}. It shows that a formal bijection between supervised learning and IRL exists. Among the covered algorithms, the \textit{Cascaded Supervised approach to IRL} (CSI) \citep{klein2013cascaded} is of particular interest to us. We refer the reader to these papers for more details and we explain in details here how the CSI paradigm can be readily applied to our setting.

The demonstrator's policy, $\pi_E$, is assumed optimal for $\R_E+\B_E$ (which is unknown), so $$\pi_E = \pi_{\R_E+\B_E}^*.$$
Write $Q_E(h,a) = Q^*_{\R_E+\B_E}(h,a)$ the associated optimal $Q$-function. It satisfies the Bellman optimality equation (writing $h=(\dots, s)$, that is $s$ the last state of the trajectory $h$ and $h'=(h, a, s')$):
\begin{equation}
\begin{split}
    Q_E(h,a) &= \R_E(s,a) +\B_E(h,a) + \gamma \mathbb{E}_{s'|s,a}[\max_{a'}Q_E(h',a')]\\
    &= \R_E(s,a)+\B_E(h,a) + \gamma \mathbb{E}_{s'|s,a}[Q_E(h',\pi_E(h'))].
\end{split}
\end{equation}
Would the optimal policy and $Q$-function be known, we could use them to recover the optimized bonus-augmented reward using this Bellman equation:
\begin{equation}
\begin{split}
    &\R_E(s,a)+\B_E(h,a) = Q_E(h,a) - \gamma \mathbb{E}_{s'|s,a}[Q_E(h',\pi_E(h'))].
\end{split}
\end{equation}
Now, the quantities in the right hand side are unknown, but they can be estimated in an indirect way.

Assuming that the actions are discrete, we can learn the policy $\pi_E$ by mapping histories to actions (using, for example, an LSTM network). Write $\hat{\pi}: \Memory \to \States$ the resulting policy, or classifier. If we train it by minimizing a cross-entropy loss, what we learn indeed are logits $\hat{Q}(h,a)$, and the classifier is $\hat{\pi}(h) = \argmax_{a} \hat{Q}(h,a)$. Said otherwise, $\hat{\pi}$ is greedy with respect to $\hat{Q}$, that can thus be interpreted as an optimal Q-function for an unknown reward. Using the Bellman equation, we can recover this reward:
\begin{equation}
    \R(s,a)+\hat{\B}(h,a) = \hat{Q}(h,a) - \gamma \mathbb{E}_{s'|s,a}[\hat{Q}(h',\hat{\pi}(h'))]. \label{eq:recovered_reward}
\end{equation}
By Bellman, as $\hat{\pi}$ is greedy w.r.t. $\hat{Q}$, we have that $\hat{\pi}$ and $\hat{Q}$ are respectively the optimal policy and $Q$-function for the reward $\R+\hat{\B}$. We cannot use directly Eq.~\eqref{eq:recovered_reward}, the model being unknown, but we can sample the right hand side and estimate $\hat{\B}$ by solving a regression problem.

Therefore, we have reduced our initial problem to a sequence of supervised learning problem. Our algorithm is indeed CSI, up to the fact that we consider trajectories instead of states, and parameterize the bonus with the reward task. As such, the theoretical results of \citet{klein2013cascaded} applies to our setting. Notably, we would have that
\begin{equation}
    0 \leq \mathbb{E}_{h\sim\pi_E}[\max_a Q^*_{R+\hat{B}}(h,a) - Q^{\pi_E}_{R+\hat{B}}(h,\pi_E(h)] \leq \mathcal{O}\left(\frac{\epsilon_1 + \epsilon_2}{1-\gamma}\right),
\end{equation}
with $\epsilon_1$ the classification error and $\epsilon_2$ the regression error. This means that the demonstrator policy is close to optimal if these errors are small, for the learnt bonus function. One could argue that this bound trivially holds for $R+\hat{B}=0$, when all behaviors are optimal. Yet, this is unlikely, as for having the classification error $\epsilon_1$ small, we must have $\hat{Q}(h,\pi_E(h)) > \hat{Q}(h,a\neq\pi_E(h))$ w.h.p., and thus learn an informative bonus.

\begin{figure}[h]%
    \centering
    \begin{minipage}{\columnwidth}
        \centering
        \includegraphics[width=1.\columnwidth]{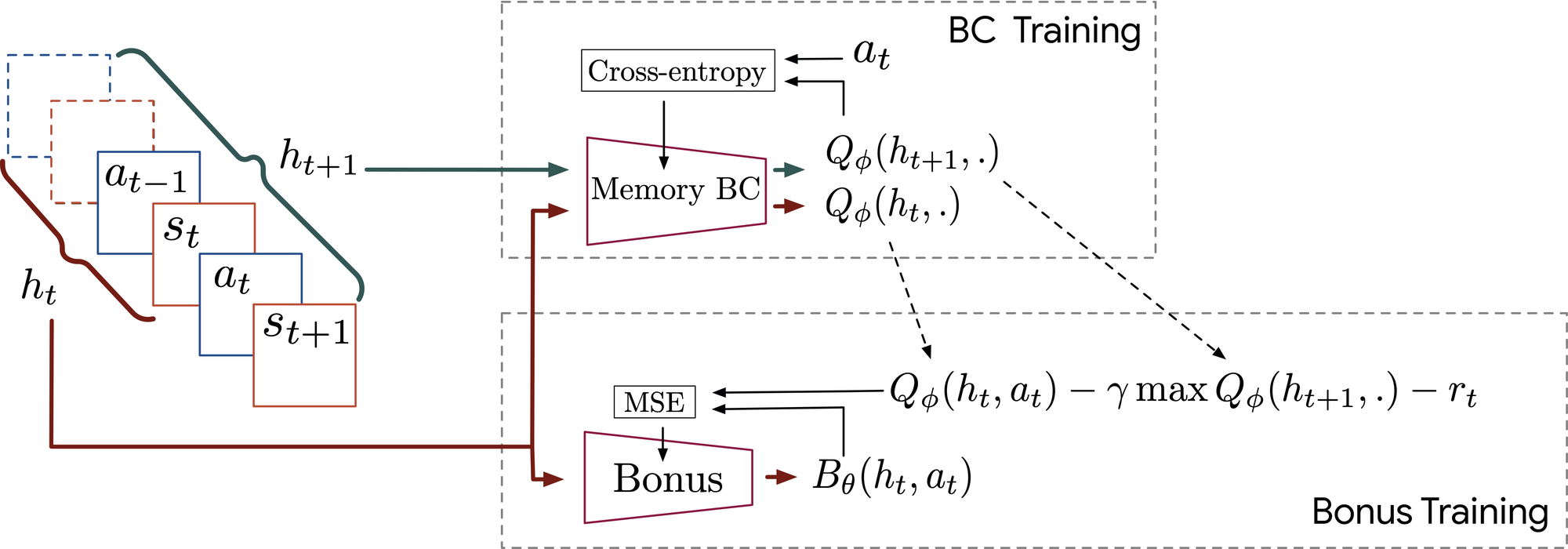}
    \end{minipage}%
\caption{Trajectories $(s,a,\dots)$ are generated by a demonstrator exploring its environment. In order to recover a bonus that can explain its behavior, a BC policy parameterized with an LSTM is trained to predict the actions of the demonstrator from its trajectories of states, by minimizing $\mathcal{L}^\text{BC}$. The policy's logits $Q_{\phi}$ are interpreted as optimal $Q$-values and used to compute a regression target. A bonus function $\B_\theta$, parameterized with an LSTM, is then trained to predict it, by minimizing $\mathcal{L}^\text{reg}$.}
\label{schema}
\end{figure}
\textbf{Implementation.} More concretely, we consider  $\sm(Q_\phi)$ to be a neural network classifier with LSTM \citep{hochreiter1997long} units, $\phi$ being the set of parameters and $Q_\phi$ being the logits. We train $\pi_\phi$ to do behavioral cloning, that is to predict the demonstrator actions $a_E$ based on its past interactions $h_E$, by minimizing a cross-entropy loss:
\begin{equation}
    \mathcal{L}^\text{BC} = - \ln(\sm(Q_\phi(h_E, a_E)),
\end{equation}
with $Q_\phi(h,a)$ the $a^\text{th}$ logit for input $h$. If the classifier learns correctly, the logits of the resulting network should satisfy $Q_\phi(h_E, a_E)>Q_\phi(h_E, a)$ for $a\neq a_E$, and the class predicted by the classifier will be $\pi_\phi(h) = \argmax_a Q_\phi(h,a)$. Hence, as explained above, one can interpret $Q_\phi$ as an optimal $Q$-function (hence the notation), and $\pi_\phi$ as the associated optimal policy. Recall that these quantities can be related to the bonus-augmented reward through  Bellman:
\begin{equation}
    Q_\phi(h,a) = \R(s,a) + \B(h,a) + \mathbb{E}_{s'|s,a}[ Q_\phi(h', \pi_\phi(h'))].
\end{equation}
Then, we can learn a network $\B_\theta$ (parameterized by $\theta$, with LSTM units) by minimizing a square-loss, the regression target being $Q_\phi(h_E, a_E) - \gamma Q_\phi({h_E}', \pi_\phi({h_E}')) - R(s_E,a_E)$, an unbiased sample of what would give the true Bellman equation. However, we only observe optimal actions (according to $\R + \B$), so this alone would hardly generalize to suboptimal ones. Therefore, we propose a heuristic, that consists in regressing for suboptimal actions towards $\B_\text{min}$, a hyperparameter of the algorithm. For example, it could be set to $\min (Q_\phi(h_E, a_E) - \gamma Q_\phi({h_E}', \pi_\phi({h_E}')) - R(s_E,a_E))$, the minimum being over transitions in the dataset. This gives the following loss, for a transition $(h_E,a_E,{h_E}')$, and for $\overbar{a_E}$ being sampled randomly in $\Actions\setminus\{a_E\}$:
\begin{equation}
\begin{split}
    \mathcal{L}^\text{reg} &= \Bigl(
    Q_\phi(h_E, a_E) - \gamma Q_\phi({h_E}', \pi_\phi({h_E}')) - R(s_E,a_E)
    - \B_\theta(h_E,a_E) \Bigr)^2 \\
    &+ \Bigl(\B_\text{min} - \B_\theta(h_E,\overbar{a_E}) \Bigr)^2.
\end{split}
\end{equation}
To sum up, we train a BC policy by minimizing $\mathcal{L}^\text{BC}$. The implicit resulting logits are considered optimal $Q$-values, that are in turn used to learn the bonus $B_\theta$ by minimizing the loss $\mathcal{L}^\text{reg}$ (Figure~\ref{schema}).

\section{Experiments}
\begin{figure}[!tbh]
    \centering
    \begin{minipage}{\columnwidth}
        \centering
        \includegraphics[width=.85\linewidth]{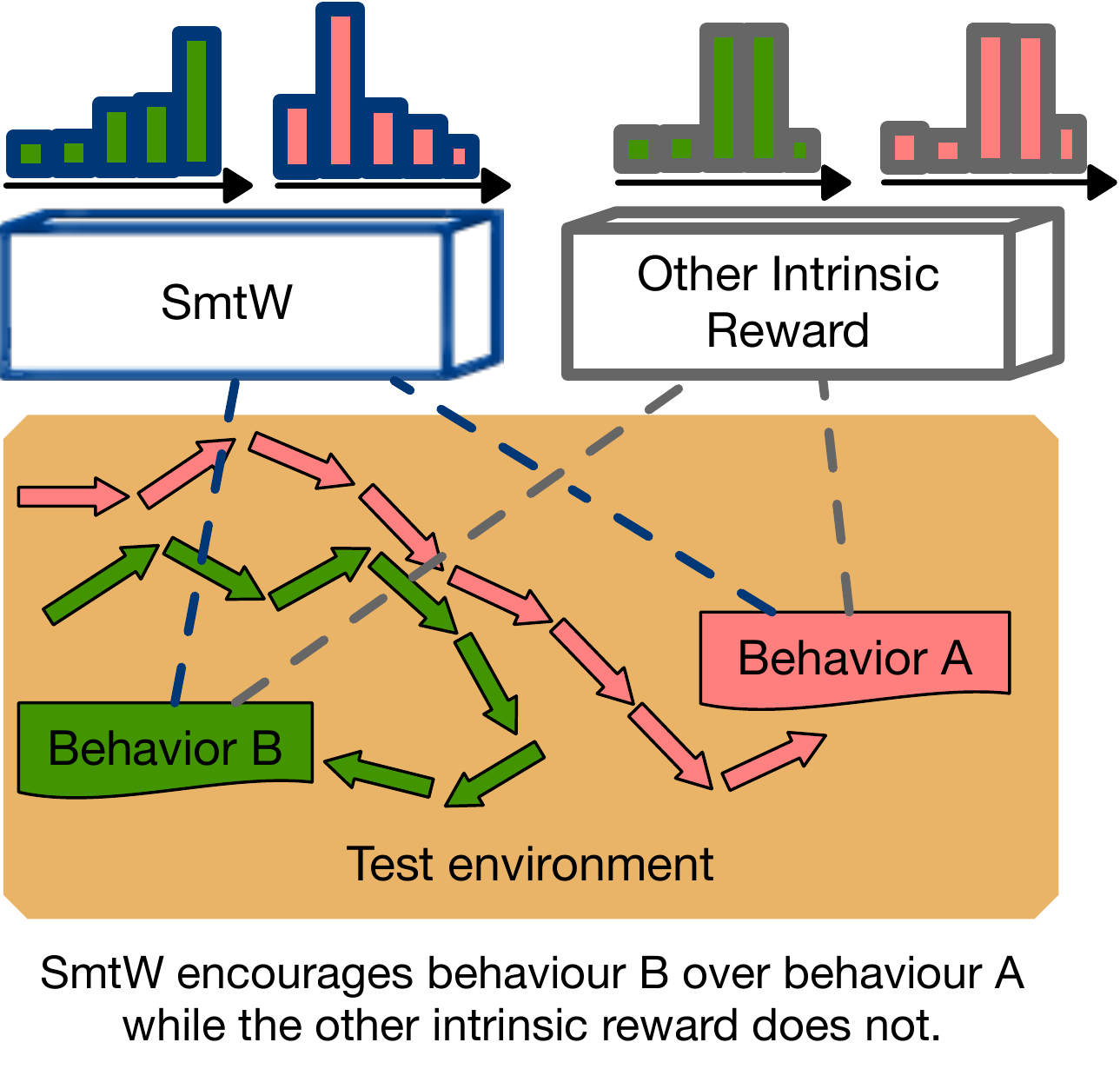}
    \end{minipage}%
\caption{Comparing intrinsic rewards on what behavior they encourage or discourage on new unseen environments.}
\label{xp_explained}
\end{figure}

We aim at providing insights on what \textit{priors} \method{} is able to extract from the demonstrations and specifically, we wish to verify that \method{} is able to encourage a \textit{structured exploration} of the environment. In order to thoroughly study the method, we test it on a grid-world where we are able to design controllers with specific behaviors.
As in IRL, studying the return of an agent trained with our bonus is only a proxy to evaluate \method{}'s quality and is not informative on the priors the bonus conveys. We thus focus our experiments on analyzing the priors that were extracted from the demonstrations by the method.
More specifically we wish to answer the following questions: (1) Is \method{} encouraging the demonstrator's behavior more than a random one? (2) Is \method{} capturing the demonstrator's style, its way of exploring the environment? (3) Does \method{} capture the skills required to solve the task? (4) Does \method{} encourage novelty seeking? (5) Does \method{} capture the constraints the demonstrator may be submitted to?
To do so, we design controlled behaviors and study the bonus returned along these specific behaviors by \method{}, as described in Fig.~\ref{xp_explained}. Given a behavior $A$ and a behavior $B$, this allows to check if a given bonus encourages behavior $A$ over $B$ or vice versa or rewards them equivalently.

After addressing these questions, we also verify that a simple agent can benefit from \method{} to actually solve efficiently a task.

\textbf{The environment.}
We introduce a specific environment to answer these. We require this environment to be procedurally-generated in order to test \method{}'s ability to generalize to unseen environments. We require the environment to be complex enough so that exhaustive exploration is prohibitively expensive. To achieve this, we introduce the KeysDoors grid-world of size NxN.
\begin{wrapfigure}{r}{0.44\columnwidth}
    \centering
    \includegraphics[width=\linewidth]{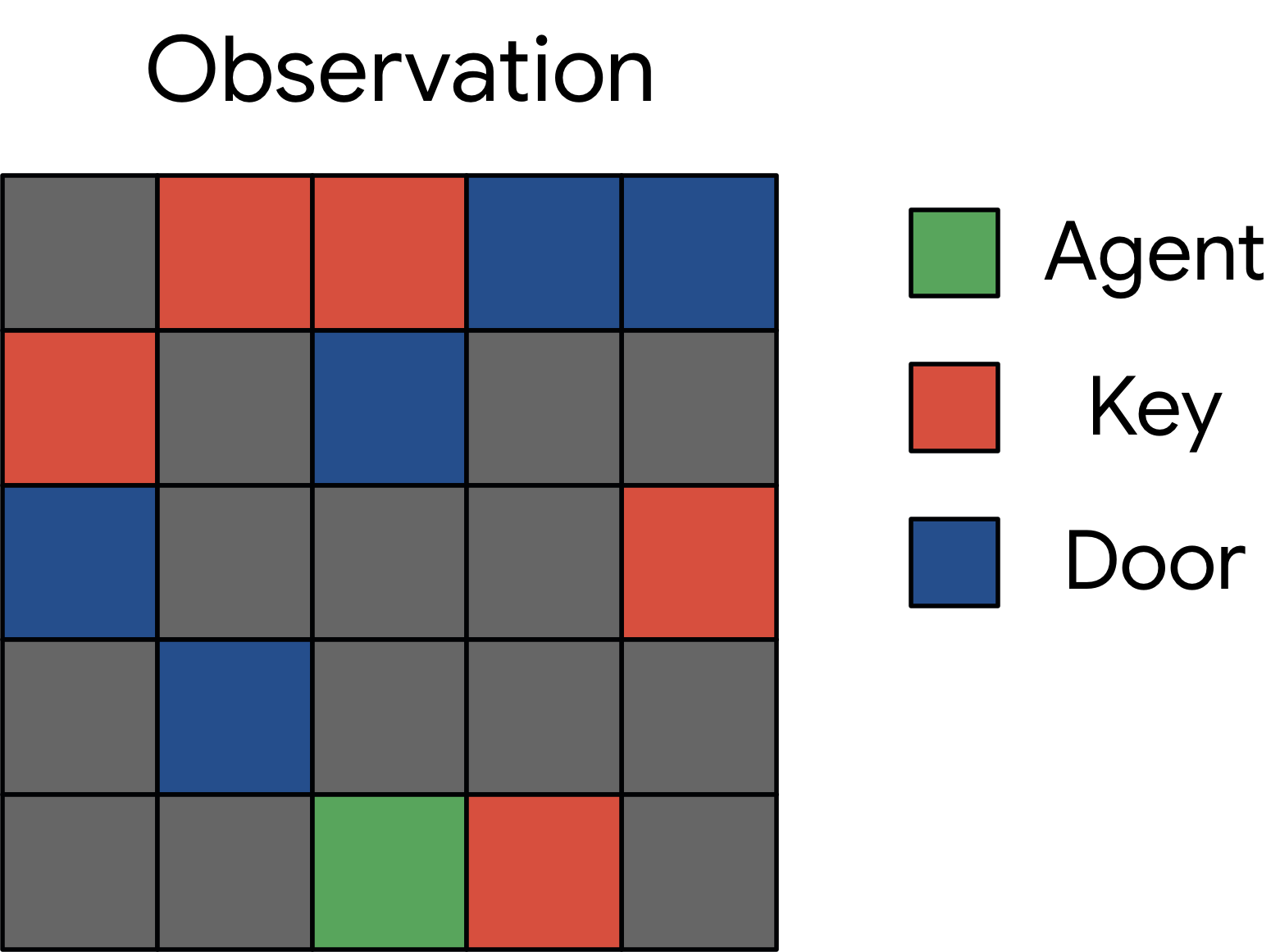}
\caption{KeysDoors(N=5).}
\label{keysdoors}
\end{wrapfigure}
It contains $N$ keys and $N$ doors, modeled by two different colors. The agent has a third color. The goal is to find the correct key and to open the correct door with it. As doors (resp. keys) are indistinguishable (except by their locations), an explorer has to try the different keys on the different doors. Actions available are \{\textit{go left, go right, go up, go down, take, open, wait}\}.  When an agent makes the action ``take'' on a key, it is then able to move with it. Actions ``open'' or ``take'' make the agent lose the key it was previously holding. To solve the task, the agent has to go to the correct key, take it, go to the correct door without doing action ``take'' or ``open'' on the way (so as not to lose the key), and then ``open'' the door. We need the environment to require \textit{perseverance} so we made the reward function -1 for any actions but the \textit{wait} action, that is rewarded 0. Opening the correct door with the correct key gives a reward of 100 and terminates the episode. It requires perseverance as a ``lazy'' policy would get a return of 0 whereas trying to find the 100 reward gives -1 at each step. This is a well known issue in RL that simple exploration leads to such lazy solutions.

\begin{figure}[!htb]
    \centering
    \includegraphics[width=\columnwidth]{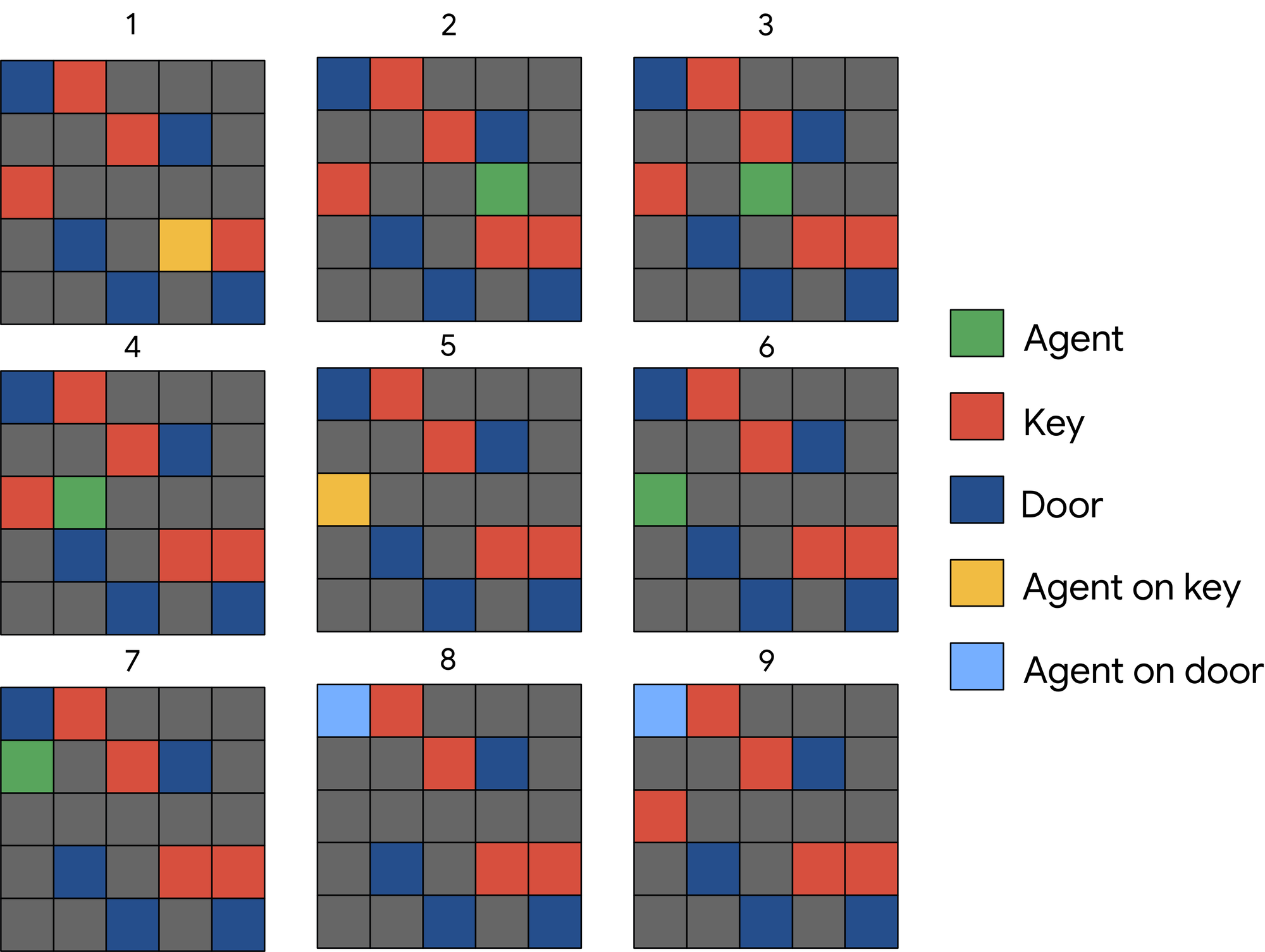}
\caption{A trajectory of length 9 in an instance of the KeysDoors(N=5) environment.}
\label{keysdoors_traj}
\end{figure}

The KeysDoors environment is generated procedurally. For each column, locations for a door and a key are sampled uniformly without replacement. Thus, there is exactly one key and one door on each column and these cannot be at the same location. The ``correct'' key is then uniformly sampled among the keys and the "correct" door is sampled uniformly among the doors. The initial position of the agent is sample uniformly on the grid. The environment gives both a ground-truth-state (an integer representing the current state), unused by \method{} as well as an RGB observation (as shown in Fig.~\ref{keysdoors}), used by \method{}.
Figure~\ref{keysdoors_traj} shows a trajectory in one possible instance of the KeysDoors environment with $N=5$. Every observation $x$ (an $N \times N \times 3$ tensor) is normalized between 0 and 1 by dividing by 255.

\textbf{The demonstrations}. For a given instance of the environment, the demonstrator navigates between keys and doors and tries key/door pairs in a precise order. It takes the first key on the left and tries it on the first door on the left, then it tries the same key on the second door etc. Once it has tried the first key on every door, it repeats the operation with the second key and proceeds further this way. The episode ends when the demonstrator finds the right key/door pair and obtains the reward. Then it ``exploits'', taking the correct key and opening directly the correct door five consecutive times. Note that this also simulates the non-stationnarity happening in most goal-directed task solving processes. One first mainly explores and then exploits more and more. 

\textbf{Train vs. Test.} The bonus is always used in new test environments, unseen in the demonstrations. \method{}'s ability to generalize to new environments is thus tested in all the following experiments. Given the possible positions of the keys, of the doors and then of the correct key and the correct door, there are $(N-1)N^3$ possible instances of the environment.
\textbf{The behaviors} that are designed to study what is actually encouraged or discouraged by \method{} are the following. Their associated bonus is always studied in test instances of the environment.
\begin{itemize}
\item The demonstrator behavior acts as described previously, sequentially trying key/door pairs.
\item The \textit{random} behavior takes random actions. Trajectories are limited to 1000 steps.
\item The \textit{demonstrator inverse} behavior is similar to the demonstrator as it navigates to a key, takes it, navigates to a door and opens it. However, the key/door pairs are tried in the reverse order to the demonstrations.
\item The \textit{demonstrator random} behavior is also similar but tries the key/door pairs in a random order.
\item The \textit{dummy demonstrator} behavior navigates exactly like the demonstrator but drops the key at a random time on the way to the door (uniformly sampled on the path to the door) by taking action \textit{open}. The trajectories are limited to 1000 steps.
\item The \textit{standing still} behavior remains in its original position by only taking the \textit{wait} action.
\item The \textit{waiting demonstrator} behavior acts like the demonstrator but has a probability 0.1 of waiting at each step.
\item The \textit{unsafe demonstrator} acts like the demonstrator but takes this action \textit{take} each time it moves until it has a key. Taking action \textit{take} somewhere else than on a key can be considered as breaking a safety constraint that the demonstrator respects strictly. 
\end{itemize}

A trajectory of an agent moving to a key, taking it, moving to a door and trying to open it with the key is shown in Fig.~\ref{keysdoors_traj}

\subsection{Bonus analysis}\label{bonus_analysis}
We train the \method{} bonus on 200 KeysDoors(N=5) training environments with 10 demonstrations for each of them. The implementation choices are detailed in Sec.~\ref{implementation_details}.
In order to study the priors extracted from the demonstrations, we study the bonus given by \method{} along various trajectories following a given behavior. 
We thereafter plot the distribution of received bonus along the various controlled behaviors on 20 test environments, unseen during the training of \method{}.
We compare the bonus given by \method{} along these trajectories to the one that would be given by a count-based \citep{strehl2008analysis} and a random network distillation bonus \citep{burda2018exploration}. The very same trajectories are presented to each bonus.
\begin{figure*}[!tbh]
    \centering
    \begin{minipage}{\textwidth}
        \centering
        \includegraphics[width=\linewidth]{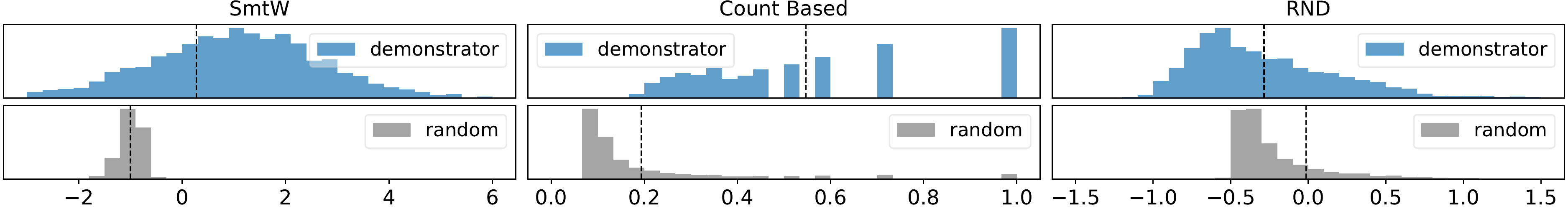}
    \end{minipage}%
\caption{Bonus distribution received by an demonstrator's behavior (top) and a random behavior (bottom). We can say that a bonus encourages more a behavior A than a behavior B if the distribution of bonus along trajectories following A are globally higher than the one along trajectories following B. \method{} encourages the demonstrator behavior over the random behavior.}
\label{hist_random}
\end{figure*}
\begin{figure*}[!tbh]
    \centering
    \begin{minipage}{\textwidth}
        \centering
        \includegraphics[width=\linewidth]{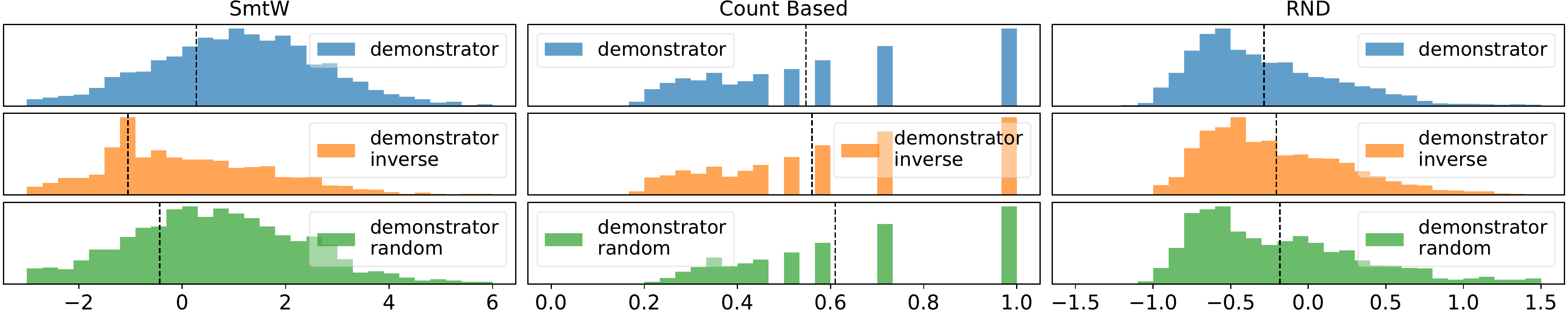}
    \end{minipage}%
\caption{Bonus distribution received by the \textit{demonstrator} behavior (top), the \textit{inverse demonstrator} behavior (middle), and the \textit{random demonstrator} behavior. (bottom).}
\label{hist_order}
\end{figure*}
\begin{figure*}[!tbh]
    \centering
    \begin{minipage}{\textwidth}
        \centering
        \includegraphics[width=\linewidth]{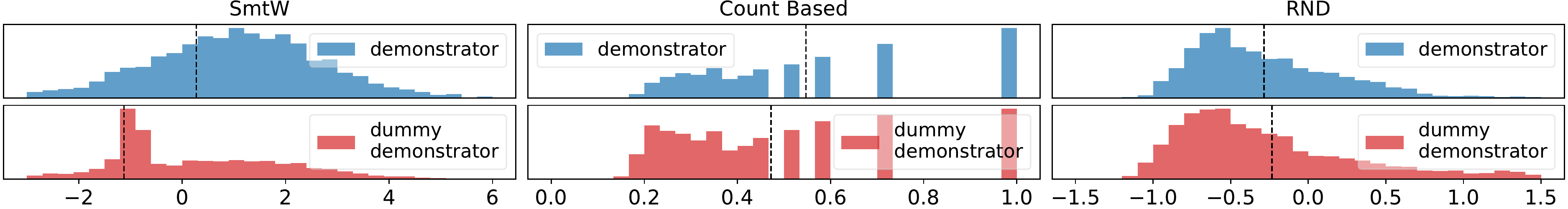}
    \end{minipage}%
\caption{Bonus distribution received by the \textit{demonstrator} behavior (top) and by the \textit{dummy demonstrator} behavior, acting (almost) like the demonstrator but releasing the key on the way to the door (bottom).}
\label{hist_releasing}
\end{figure*}

\textbf{Does \method{} encourage a structured exploration more than a random one?} We compare in Figure~\ref{hist_random} the distribution of bonus received along random trajectories to the ones obtained by the demonstrator's behavior. Recall that \method{} has been trained on similar environments but is here tested on different ones. It is thus provided with trajectories unseen during training.

As shown on Figure~\ref{hist_random}, the demonstrator's behavior (top) is more rewarded by \method{} than the random behavior (bottom).
We can conclude that \method{} encourage the agent to follow a demonstrator-like behavior on new unseen environments more than a random behavior.
The count-based bonus also rewards the demonstrator's behavior more than a random one as a random behavior explores the environment very locally. Surprisingly, RND rewards the random behavior more than the demonstrator's one. This might be explained by the fact that the demonstrator visits several times the same state in order to explore correctly. Indeed the demonstrator has to go several times to the same key to take it and try it on the several doors.

\textbf{Does \method{} capture the demonstrator's style, his way of exploring the environment?} We show in Figure~\ref{hist_order} the distribution of bonus received along different behaviors: the \textit{demonstrator} one, the \textit{demonstrator inverse} one as well as the \textit{demonstrator random} one. These three behaviors lead to the same outcome but we hope to capture the demonstrator's exploration bias and see if it encourages the behaviors that tries the key/door pair in the same order as in the demonstrations.
As shown on Figure~\ref{hist_order}, the count-based bonus and RND reward similarly the three behaviors, as they lead to the same amount of novelty. Only the order in which the key/door pairs are tried is change. \method{}, on the contrary, encourages to reproduce the demonstrator bias. It rewards more the  behavior trying the key/door pairs in the same order as in the demonstrations.

\textbf{Does \method{} capture the priors useful to solve the task?} Figure~\ref{hist_releasing} shows the distribution of bonus received by the \textit{demonstrator} behavior and compares the bonus received to the one received when following the \textit{dummy demonstrator} one. 
As shown on Figure~\ref{hist_releasing}, the count-based bonus and RND reward equivalently these two behaviors as they bring the same amount of novelty (both in term of ground-truth-state and observations). \method{} does not reward the \textit{dummy demonstrator} behavior as much as the expert one and we can interpret the lower distribution mode (\method{}-bottom) as the bonus obtained after loosing the key. We can argue that \method{} has somehow captured the prior that it is useful to navigate from the key to the door without loosing the key, as it rewards more the \textit{demonstrator} behavior than the \textit{dummy demonstrator} one.
\begin{figure*}[!tbh]
    \centering
    \begin{minipage}{\textwidth}
        \centering
        \includegraphics[width=\linewidth]{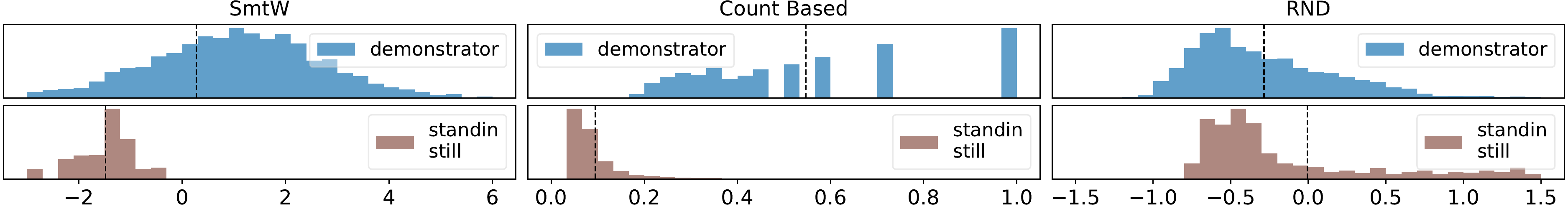}
    \end{minipage}%
\caption{Bonus distribution received by the \textit{demonstrator} behavior (top) and by the \textit{standing still} behavior (bottom).}
\label{hist_novelty}
\end{figure*}
\begin{figure*}[!tbh]
    \centering
    \begin{minipage}{\textwidth}
        \centering
        \includegraphics[width=\linewidth]{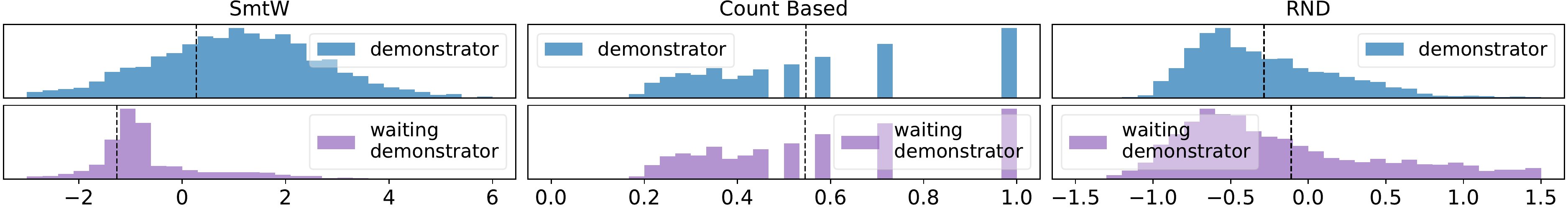}
    \end{minipage}%
\caption{Bonus distribution received by the \textit{demonstrator} behavior (top) and by the \textit{waiting demonstrator} behavior (bottom).}
\label{hist_time}
\end{figure*}
\begin{figure*}[!tbh]
    \centering
    \begin{minipage}{\textwidth}
        \centering
        \includegraphics[width=\linewidth]{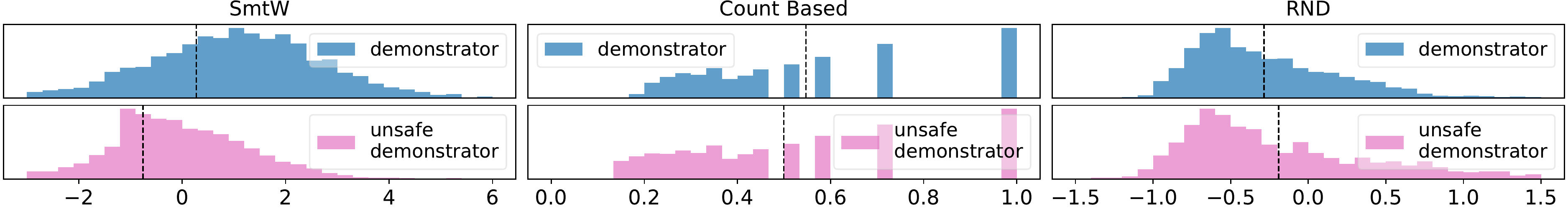}
    \end{minipage}%
\caption{Bonus distribution received by the \textit{demonstrator} behavior (top) vs. by the \textit{unsafe  demonstrator} behavior (bottom).}
\label{hist_safety}
\end{figure*}

\textbf{Does \method{} encourage long-term exploration?} As the environment gives a reward of $-1$ for taking any action but the \textit{wait} action, an agent not exploring sufficiently would quickly converge to the policy only taking action \textit{wait} to avoid negative rewards (verified in Figure~\ref{res}). This same problem is visible in the \textit{Pitfall!} game, where the best agents learn a policy obtaining 0 reward, while persevering humans get much higher scores. We show in Figure~\ref{hist_novelty} the distribution of bonus obtained by the \textit{standing still} behavior.
As shown on Figure~\ref{hist_novelty}, \method{} rewards much less a behavior not seeking novelty. As expected the count based gives a bonus very close to 0 for such a behavior. Perhaps surprisingly, RND rewards negatively this behavior but not with an average bonus lower than the demonstrator's behavior. This might be also due to the designed bonus normalization that RND uses (zero-mean unit-variance). 

\textbf{Does \method{} capture the constraints the demonstrator may be submitted to?} A demonstrator can be subject to time or energy constraints. In the demonstrations, the demonstrator tries to explore the environment as fast as possible and does not take action \textit{wait} on his way to keys and doors. We compare the bonus distribution obtained by the \textit{waiting demonstrator} behavior to the one obtained by the \textit{demonstrator} one.

As shown on Figure~\ref{hist_time}, RND and the count-based bonus reward equivalently these two behaviors. On the other hand, \method{} rewards less the \textit{waiting demonstrator} behavior. We argue it has somehow captured the prior resulting from the resource constraint that leads the demonstrator to try the key/door pairs as fast as possible. In other words, it favors behaviors that, as shown in the demonstrations, discard the \textit{wait} action to simplify exploration of the MDP.
What is more, a demonstrator might be subject to safety constraints. As example, it might be dangerous for a robot to try an action in an inappropriate place. The demonstrations minimize the number of time they use the action ``take'' and only do it when on keys. We can consider that the demonstrator's behavior complied with safety constraints. We show in Figure~\ref{hist_safety} the bonus distribution obtained by the demonstrator's behavior and compare it with the one obtained by the \textit{unsafe demonstrator}.
As shown on Figure~\ref{hist_safety}, the RND and the count-based bonuses reward equivalently these two behaviors. This is expected as they bring the same amount of novelty. In contrast, \method{} rewards less the \textit{unsafe demonstrator} behavior, capturing the safety prior the demonstrator have been subject to.

Overall, we argue that \method{} is able to recover some important bias and constraints inherent to the demonstrations. Hand-crafting a bonus expressing these motivations could be extremely complicated and we demonstrated that \method{} is able to generalize these motivations to unseen environments. 
\subsection{Training an agent on the bonus}
We now wish to check that an agent can benefit from \method{}. We thus train a $Q$-learning agent with \method{} and compare the results with that of a simple $\epsilon$-greedy ($\epsilon$=0.1) exploration strategy and a count-based bonus with $B(s, a) =  N(s, a)^{-1/2}$.

\begin{figure}[!h]%
    \includegraphics[width=\columnwidth]{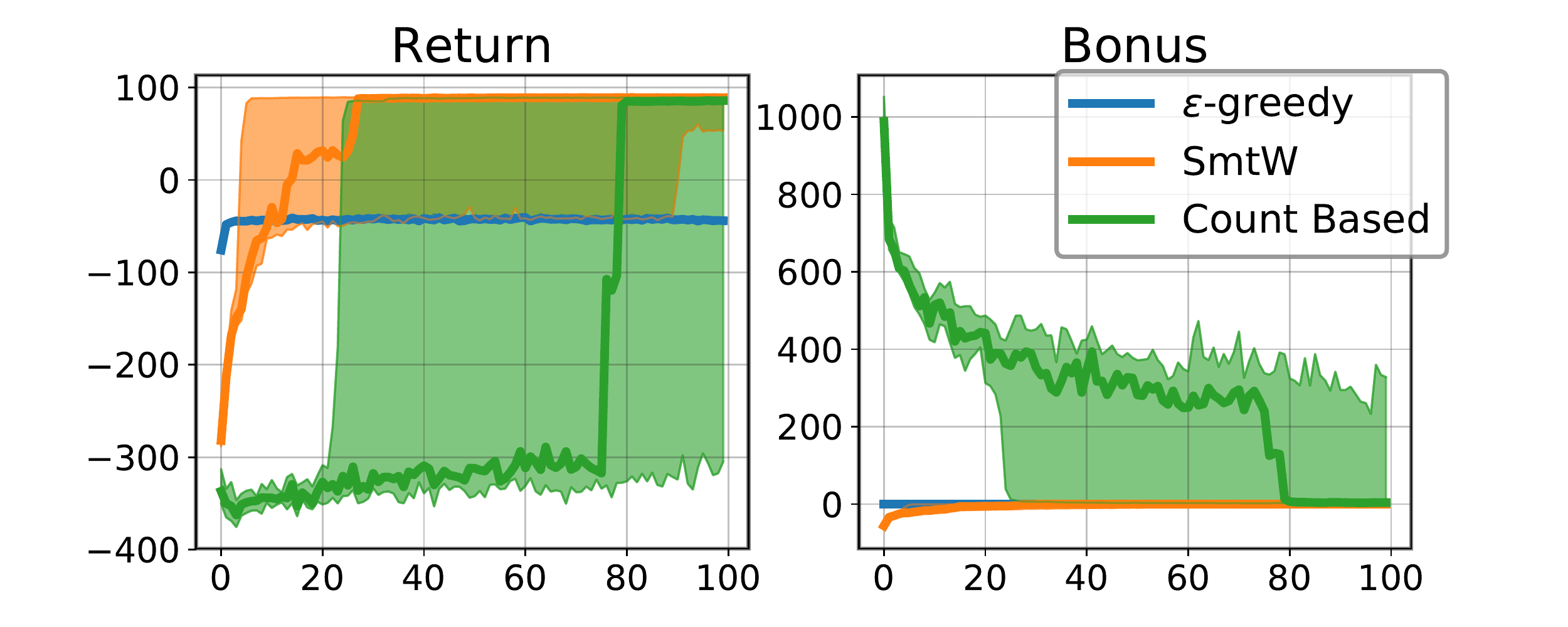}
\caption{Median and min/max values of the return per episode (left) and of the total bonus per episode(right).}
\vspace{-10pt}
\label{res}
\end{figure}

The results are averaged over 10 newly generated environments, unseen during \method{} training. 
For each of these environments, the experiment is repeated twice. We present, for each algorithm, the best result after a hyper-parameter search.
The bonus given by our method is computed to capture the exploratory behavior of the demonstrator. In order for the agent not to keep exploring forever, our bonus is here divided by $\sqrt{k}$ with k the number of step of training.

As Figure~\ref{res} shows, the Q-learning with an $\epsilon$-greedy exploration strategy quickly gets stuck in  ``waiting'' at each timestep. \method{} encourages the agent to visit its environment and solves the 10 new environments much faster than the count-based method that push for exhaustive exploration.

\subsection{Implementation Details}
\label{implementation_details}
Our method works directly with visual inputs, as shown in Fig.~\ref{keysdoors}.
The network used for the behavioral cloning policy $\pi_{\phi}$ has the following architecture: an LSTM with 64 units, a fully-connected layer with 512 units and relu activation and an output layer with as many units as there are actions available in the environment (7 for KeysDoors).
It is trained with the Adam optimizer~\citep{kingma2014adam} with a learning rate of $10^{-3}$ and a batch size of $1$. It uses the visual input from the environment and not the ground-truth state.

The network used for the regression of the bonus $\B_\theta$ has the same architecture but an output layer with a single unit.
It is trained with the Adam optimizer, a learning rate of $10^{-4}$ and a batch size of $1$.
The discount factor used in \method{} is set to $\gamma=0.99$.

For experiment shown in Figure~\ref{res}, the tabular $Q$-learning is trained on the 10 test environments twice and the figure shows the median and the min/max values. For each of the compared algorithms, we sweep over the agent learning rate over the following values: $[0.01, 0.1, 0.5, 0.7]$. Only the result of the learning rate with the highest median return over the $10 \times 2$ runs is shown for each algorithm. The $\varepsilon$-greedy strategy is used for all methods with $\varepsilon=0.1$.
Even though the agent is tabular, we recall that $\method{}$ itself does not access the ground-truth state of the environment. It works from observations. The count-based bonus, on the contrary, counts ground-truth states.

\section{Related Work}
\textbf{Intrinsic Motivation.} Intrinsic motivation is essential to mental development \citep{oudeyer2007intrinsic} and we can argue that this may, in consequence, be an essential component for computational learning. \citet{oudeyer2009intrinsic} argue that all humans respond to intrinsic motivations. Young infants motivations can be qualified as more chaotic as they push children to bite, throw, grasp or shout in order to learn. Adults, in contrast, have more structured intrinsic motivations, activated, for instance, when they play games, read novels or watch movies. Correctly using these numerous intrinsic motivations can be key to train agents that solve more and more difficult tasks.
Instead of modeling such intrinsic motivations to mimic cognitive processes, we learn them from demonstrations.

\textbf{Exploration.} In order to provide an exploration signal to the agent,~ \citep{strehl2008analysis} proposed the very intuitive count-based method in order to measure novelty. Counting how many times the agent has been in a given state, it rewards less visited states. Several methods extended this idea to large state-space problems~ \citep{ostrovski2017count, bellemare2016unifying, tang2017exploration, machado2018revisiting}, where it is not possible to count state occupancy. Intrinsic curiosity is also commonly computed as a prediction error, either trying to predict the environment's dynamics~ \citep{pathak2017curiosity, raileanu2020ride} or random statistics about the current state~ \citep{burda2018exploration}. Different methods try also to measure surprise as a prediction gain~ \citep{schmidhuber1991curious, houthooft2016variational}. Instead of designing such a bonus, we aim at learning one from demonstrations.

\textbf{Learning from demonstrations.} Imitation learning, the problem of learning from demonstrations, is typically folded into two different paradigms. (1) Behavioral cloning~\citep{pomerleau1991efficient, bagnell2007boosting, ross2010efficient} tries to directly match the demonstrator's behavior, generally using supervised learning techniques. (2) Inverse Reinforcement Learning ~\citep{russell1998learning, ng2000algorithms} first tries to recover a reward explaining the demonstrator's behavior, before optimizing the reward for imitating the demonstrator. Some methods output an explicit reward~ \citep{klein2013cascaded, abbeel2004apprenticeship, ng2000algorithms, ziebart2008maximum} while adversarial imitation learning can be seen as IRL with implicit reward recovery~ \citep{ho2016generative,fu2017learning,finn2016guided}. Overall these methods all assume that the near-optimality of the demonstrations. Some works try to relax this assumption and to learn from sub-optimal demonstrations~ \citep{jacq2019learning, brown2019extrapolating}. IRL methods typically control the quality of their algorithm through the proxy of the return obtained by an agent trained on the inferred reward. 

Our methods differs from these methods it does not assume that demonstrations are optimal but rather try to answer  the question: ``In what way is the demonstrator's behavior deviating from an optimal policy?''.
Moreover, we do not seek to recover a reward as in IRL but rather to recover a bonus explaining which, added to the environment reward, explains the demonstrator's behavior. Facing the same problem that the usual proxy to control the algorithm quality (training an agent on the inferred bonus) is not informative, we decided to study our method through its response to various behaviors.
\section{Conclusion}
In this work, we present a novel method for extracting an intrinsic bonus from the demonstrations. The method we introduce is offline and does not require environment interactions to recover the bonus, unlike recent adversarial imitation methods who need numerous interaction in order to recover a reward function. Anyway, those methods could not be readily applied to our problem, as they do not explicitly compute a reward function. Moreover, to the best of our knowledge, this is one of the very first method to recover some kind of reward that is history-dependent. We show how this bonus generalizes to unseen environments and is able to convey long-term priors. We exemplified the approach on a simple yet didactic and challenging example. Yet, testing the method on a larger-scale environment would require human exploratory demonstrations. Gathering such a dataset is costly and very few are already available, none of them really covering our setting.
Even though the given example is simple, this novel approach of capturing the demonstrator's bias could potentially lead to new lines of work in RL. For instance, one could use our method to implement \textit{behavioral style-transfer} in RL and show to an agent a specific way to solve the task thanks to demonstrations. Combining a reward and biases extracted from demonstrations may also help for robotic tasks, where some aspects of the task are easily programmable with a reward but some expectations on how to solve the task may be easier to transmit thanks to demonstrations. This could also lead to some advances in tackling mispecified rewards. Using both a reward, that would contain information on the task to solve but not fully describe the constraints of the problem and demonstrations to correct the reward can be key to train sequential controllers in complex dynamics.

\newpage

\bibliographystyle{ACM-Reference-Format} 
\bibliography{sample}

\end{document}